\documentclass[letterpaper]{article} 
\usepackage{aaai2026}  
\usepackage{times}  
\usepackage{helvet}  
\usepackage{courier}  
\usepackage[hyphens]{url}  
\usepackage{graphicx} 
\usepackage{natbib}  
\usepackage{caption} 
\usepackage{algorithm}
\usepackage{algorithmic}

\definecolor{lightpink}{rgb}{0.9, 0.41, 0.56}
\usepackage{amsmath}
\usepackage{url}            
\usepackage{booktabs}       
\usepackage{amsfonts}       
\usepackage{nicefrac}       
\usepackage{microtype}      
\usepackage{xpatch}
\usepackage{graphicx}
\usepackage{epstopdf}
\usepackage{multirow}
\usepackage{xpatch}
\usepackage{amssymb}

\DeclareMathOperator*{\argmin}{argmin}
\usepackage{newfloat}
\usepackage{listings}
\DeclareCaptionStyle{ruled}{labelfont=normalfont,labelsep=colon,strut=off} 
\floatstyle{ruled}
\newfloat{listing}{tb}{lst}{}
\floatname{listing}{Listing}
\nocopyright 

\title{Improving Deepfake Detection with Reinforcement \\ Learning-Based Adaptive Data Augmentation}
\author{
    Yuxuan Zhou\textsuperscript{\rm 1},
    Tao Yu\textsuperscript{\rm 3},
    Wen Huang\textsuperscript{\rm 1},
    Yuheng Zhang\textsuperscript{\rm 1},
    Tao Dai\textsuperscript{\rm 2}\thanks{Corresponding Author.},
    Shu-Tao Xia\textsuperscript{\rm 1}
}
\affiliations{
    \textsuperscript{\rm 1}Tsinghua University,
    \textsuperscript{\rm 2}Shenzhen University,
    \textsuperscript{\rm 3}CASIA \\
    zhouyuxuan25@mails.tsinghua.edu.cn
}

\usepackage{bibentry}

\begin{document}

\maketitle

\begin{abstract}

The generalization capability of deepfake detectors is critical for real-world use. Data augmentation via synthetic fake face generation effectively enhances generalization, yet current SoTA methods rely on fixed strategies—raising a key question: \textit{Is a single static augmentation sufficient, or does the diversity of forgery features demand dynamic approaches?} We argue existing methods overlook the evolving complexity of real-world forgeries (e.g., facial warping, expression manipulation), which fixed policies cannot fully simulate.To address this, we propose CRDA (Curriculum Reinforcement-Learning Data Augmentation), a novel framework guiding detectors to progressively master multi-domain forgery features from simple to complex. CRDA synthesizes augmented samples via a configurable pool of forgery operations and dynamically generates adversarial samples tailored to the detector’s current learning state.Central to our approach is integrating reinforcement learning (RL) and causal inference. An RL agent dynamically selects augmentation actions based on detector performance to efficiently explore the vast augmentation space, adapting to increasingly challenging forgeries. Simultaneously, the agent introduces action space variations to generate heterogeneous forgery patterns, guided by causal inference to mitigate spurious correlations—suppressing task-irrelevant biases and focusing on causally invariant features. This integration ensures robust generalization by decoupling synthetic augmentation patterns from the model’s learned representations.Extensive experiments show our method significantly improves detector generalizability, outperforming SOTA methods across multiple cross-domain datasets. 
\end{abstract}

\section{Introduction}
Deepfake technology, which generates highly realistic visual content, has drawn significant attention but is frequently exploited for malicious purposes such as disinformation dissemination and fraud, posing severe societal harms. Thus, developing robust and reliable deepfake detection systems is imperative.

While existing deepfake detection methods perform well on standardized benchmarks, real-world deployment requires handling forgeries from diverse techniques across heterogeneous environments, making generalization the key metric for robustness. Traditional methods relying solely on raw training data lack generalizability. Each forgery technique has unique artifacts, and over-optimizing for specific manipulations leads to bias. Synthetic face generation, as an effective data augmentation strategy, partially replicates real forgery processes, preserving and amplifying manipulation artifacts. However, current methods use fixed numbers of synthetic faces and static augmentation strategies during training.

This paper investigates a critical question in deepfake detectors training: \textbf{Is a single fixed augmentation strategy sufficient, or does the diversity of forgery features necessitate dynamic adaptation?} Empirical studies \cite{Prodet} reveal that mixing synthetic samples from conflicting augmentation strategies—such as spatial-domain blurring and frequency-domain noise—often degrades performance. For instance, spatial-domain augmentations may introduce low-frequency blur that interferes with frequency forgery features, leading to suboptimal detection. Consequently, most prior work evaluates single-strategy augmentation in isolation. However, this approach contradicts the expectation that detectors trained on diverse forgery patterns should learn richer, more robust features. We identify the root cause as the absence of coordinated strategy scheduling, which induces incompatible feature conflicts across augmentation domains. Different augmentations inherently embed domain-specific artifacts (e.g., spatial warping patterns in FaceSwap \cite{fs} vs. spectral inconsistencies in FreqBlender \cite{freqblender}). Naively mixing these strategies during training will force detectors to develop competing feature preferences—oscillating between overfitting to dominant artifacts while neglecting causally stable forgery footprints. This bias becomes catastrophic when encountering unseen forgeries.

To address these limitations, we propose Curriculum Reinforcement-Learning Data Augmentation (CRDA), an innovative framework integrating curriculum learning with adversarial reinforcement learning. CRDA’s core idea is to guide the detector through a progressive learning process by designing increasingly difficult forged samples and optimizing strategies via dynamic feedback, augmented by causal learning theory to suppress task-irrelevant spurious correlations. It consists of three key components: 1) \textbf{RL-based dynamic augmentation strategy:} the agent dynamically selects optimal, challenging data augmentation methods based on real-time model state perception, optimizing via multi-objective rewards. This forces the detector to learn universal features through the continuous evolution of forged samples in an adversarial game. 2) \textbf{Multi-environment causal invariance learning:} leveraging causal inference, we analyze bias sources from data augmentation feature conflicts, introducing Invariant Risk Minimization (IRM) for training on biased data. Auxiliary environments are constructed using historical entropy from the PolicyNet. 3) \textbf{Multi-dimensional curriculum scheduling:} the curriculum strategy is extended to three decoupled dimensions—controlling the proportion of augmented synthetic faces, regulating RL exploration intensity, and adjusting forgery region scale/area—all scheduled by training phase.

Briefly, the main contributions of this work are summarized as follows:  
\begin{itemize}
    \item To our best knowledge, CRDA is the first study in deepfake detection to systematically explore and integrate the interaction mechanisms of multiple data augmentations. 
    \item We propose a strategy that combines curriculum learning and reinforcement learning to guide the detector in progressively mastering multi-domain forgery identification skills from simple to complex.  
    \item We apply causal inference theory to deepfake detection, conduct an in-depth analysis of potential bias sources from the perspective of data augmentation feature conflicts, and propose corresponding solutions.
\end{itemize}
\section{Related Works}
\subsection{Data Augmentation Towards Deepfake Detectors Generalization }
Deepfake detection \cite{deepfakebench,kaur2024deepfake,heidari2024deepfake,nguyen2024laa,ba2024exposing} techniques identify manipulated images by analyzing discrepancies between real and synthetic facial data across multiple dimensions, including identity features \cite{zhou2021joint}, spatial artifacts \cite{lips,xray,I2G}, frequency-domain differences \cite{freqblender,fsbi}, and architecture-specific patterns \cite{Two-branch,aunet}. Among various strategies, data augmentation \cite{FWA,xray,adv,Prodet,SBI,fsbi,freqblender} has become a key approach to enhancing cross-domain robustness. 
Early approaches such as FWA \cite{FWA} employed a self-blending strategy through facial region downsampling and spatial warping to simulate deepfake artifacts. Face X-ray \cite{xray} explicitly trained detectors to identify blending boundaries, while I2G \cite{I2G} extended this concept with pairwise self-consistency learning to detect intra-image inconsistencies. SLADD \cite{adv} introduced adversarial training by dynamically generating challenging blending patterns. 
More recent advancements like SBI \cite{SBI} achieve high-fidelity augmentation through intra-identity face swapping, effectively mimicking state-of-the-art manipulation techniques. FSBI \cite{fsbi} expands SBI into the frequency domain innovatively, and FreqBlender \cite{freqblender} uses a specialized frequency parsing network to synthesize fake faces by leveraging inherent correlations among frequency knowledge.
\subsection{Reinforcement Learning}
Reinforcement learning (RL) \cite{RL} is a machine learning category where an agent learns an optimal policy through trial-and-error interaction with the environment to maximize long-term cumulative rewards. It has proven effective in data augmentation in several tasks \cite{autoaugment,advautoaugment,medaugment,selfaugment}: AutoAugment \cite{autoaugment} by Google Brain uses RL to automatically search for optimal augmentation policies on the validation set, optimizing classification accuracy with strong transferability. Liu et al.\cite{medaugment} applied this to medical image segmentation, improving accuracy via Adaptive Sequence-length based Deep Reinforcement Learning. However, most RL-based augmentation schemes focus on general tasks, while ours is specifically designed for deepfake detection.
\subsection{Causal Inference}
Causal inference \cite{causal-infer} is a key AI methodology that determines if and how one variable directly affects another, and quantifies such effects. Extensive research \cite{IRM,casul2,dm-irm,ood-irm,bayesian-irm} exists on eliminating dataset bias. Jones et al.\cite{casual1} used causal inference to examine medical image biases, proposing three mechanisms for fairness. Arjovsky et al.\cite{IRM} derived feature invariance from causality and introduced Invariant Risk Minimization (IRM) to constrain models to learn stable correlations, addressing reliance on data biases. Lin et al.\cite{casul2} explored environmental partitioning without information loss. We are the first to introduce causal inference to deepfake detection, guiding the model to learn unbiased multi-domain features.


\section{Methodology}
\begin{figure*}
    \centering
    \includegraphics[width=0.9\linewidth]{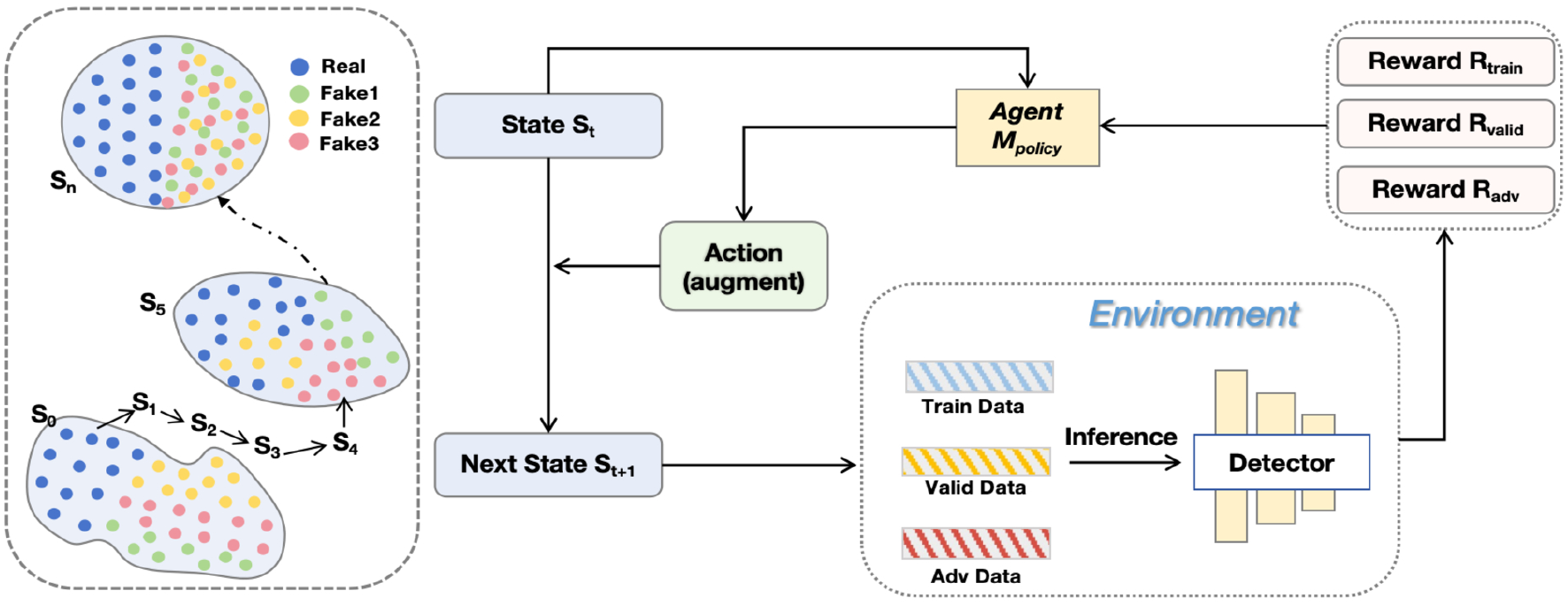}
    \caption{\textbf{PolicyNet Optimization via Multi-Rewards.} We train the RL PolicyNet using a multi-scale reward function (as shown in right panel) to guide the detector to learn features from easy to difficult. This strategy arranges different types of fake samples in a type-agnostic manner (as shown in left panel) within the detector’s latent space $S$,thereby improving its generalization.}
    \label{fig:RL}
\end{figure*}

\begin{figure*}
    \centering
    \includegraphics[width=0.92\linewidth]{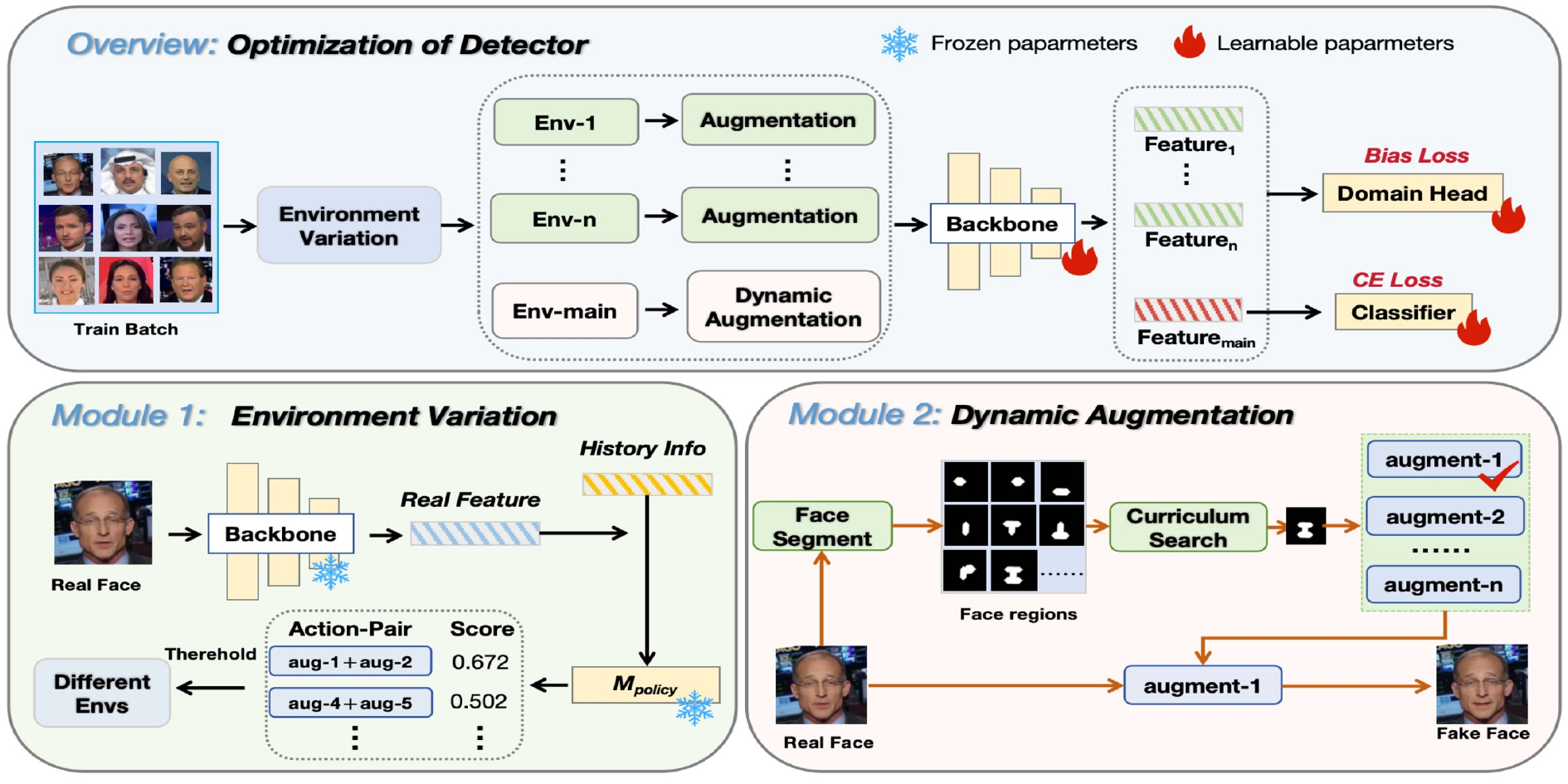}
    \caption{\textbf{Training process of CRDA detector:} In each round of optimization, CRDA generates several environments through Module 1: Environment Variation based on the PolicyNet. The main environment and auxiliary environments, after different data augmentations, optimize the detector via \(\mathcal{L}_{\text{CE}}\) and \(\mathcal{L}_{\text{bias}}\) losses respectively. In this process, the PolicyNet remains frozen.}
    \label{fig:overview}
\end{figure*}

Given the limited integration of diverse data augmentations in prior studies, our work aims to develop an effective framework capable of comprehensively and equitably learning multi-domain forgery information from various data augmentation approaches. \textbf{The overall architecture is illustrated in Figure \ref{fig:RL} for the RL PolicyNet training methodology and Figure \ref{fig:overview} for the detector training pipeline.} The following sections will provide detailed technical descriptions of our proposed methodology through its constituent modules.

\subsection{RL-based Dynamic Augmentation Strategy}
We start this section with a question: \textit{What are the advantages of our RL-based data augmentation?} Previous work CDFA\cite{fakeit} first used various data augmentation strategies to train deepfake detectors. It introduced a dynamic forgery search strategy (DFS) to optimize the strategy network, dynamically selecting forgery augmentation operations based on the validation set's performance and achieving satisfactory results. However, DFS's reliance on the validation set means the selected operations are a local optimum for that specific set. Moreover, its augmentation strategies are all variants of spatial blending operations, and the repetitive features generated limit the detector's generalization capability. In contrast, RL can explore a broader range of optimal strategies through interaction with the environment. We expanded the data augmentation operations and designed three reward functions corresponding to different levels of forgery features, using their gradual integration to guide the detector in mastering multi-domain characteristics step by step. As shown in Figure~\ref{fig:RL}, the core elements are defined as follows:

\textbf{State.} The state space of this RL is the latent space of the detector $F$. For each episode, the initial state $s_0$ is a $k$-dimensional standard normal random vector: $s_{0}\sim\mathcal{N}_k(\mathbf{0},\mathbf{1}), s_0\in\mathbb{R}^{k} $, where $k$ is the dimension of latent space. At every step $t$, the state $s_t$ is updated by the action $a_t$.

\textbf{Action.} The action space is the output space of the PolicyNet $P$, a set of data augmentation operations representing the PolicyNet's guidance on the detector $F$. In this paper, the action space $\mathcal{A} = \{a_j\}_{j=1}^7$ corresponds to seven augmentation strategies by default:
\begin{equation}
\Pi_\theta(a|s_t) = \text{Softmax}( \mathbf{W}_2 \text{ReLU}(\mathbf{W}_1 s_t + \mathbf{b}_1) + \mathbf{b}_2 )
\end{equation}
The PolicyNet uses weight matrices $\mathbf{W}_1$, $\mathbf{W}_2$ and bias terms $\mathbf{b}_1$, $\mathbf{b}_2$ to transform the latent state $s_t$ into action probabilities, with the softmax function converting outputs into probabilities for the seven strategies. These include four benchmark forgery methods from FaceForensics++ \cite{ffd}: Deepfakes \cite{df}, Face2Face \cite{f2f}, FaceSwap \cite{fs}, NeuralTextures \cite{nt}; the advanced SBI \cite{SBI} method; and frequency-domain augmentations FSBI \cite{fsbi} and FreqBlender \cite{freqblender}.

\textbf{Reward.} After state update via an action, the agent receives a reward to produce better actions. To balance training efficiency and generalization, the reward guides the model to learn basic features early, challenging features in the middle, and encourage exploration later, comprising:

1. \textbf{Training Stability Term} ($\lambda_1 \mathbb{E}[1-\mathcal{C}_{tr}]$):
$\mathbb{E}[1-\mathcal{C}_{tr}] = \frac{1}{N}\sum_{i=1}^N (1 - |f_\theta(x_i) - y_i|) $
Encourages confidence in correct predictions while preventing overfitting, aiding the detector in rapidly establishing capabilities early.

2. \textbf{Validation Performance Improvement} ($\lambda_2 \Delta \text{AUC}_{val}$):
$\Delta \text{AUC}_{val} = \text{AUC}^{(t)}_{val} - \text{AUC}^{(t-1)}_{val} $
Measures incremental AUC improvement on the validation set (composed of complex augmented fake faces) to drive generalization.

3. \textbf{Adversarial Deception Metric} ($\lambda_3 \mathcal{C}_{adv}$):
$\mathcal{C}_{adv} = \mathbb{E}_{x\sim\mathcal{D}_{real}} [ \sigma ( g_{\beta}( \sum_{j=1}^{|\mathbb{T}|} \Pi_\theta(a_j|s_t) \cdot f_\alpha(\tau_j(x))) )_{y=0}] $.
Here, $\mathbb{T} $ is the set of augmentation strategies; $\Pi_\theta(a_j|s_t)$ is the selection probability of strategy $\tau_j$; $\tau_j(x)$ applies the $j$-th operation to real sample $x$; $g_{\beta} $ is the detector's classification head; $f_{\alpha}$ extracts features from $\tau_j(x)$. Weighted summed features input to $g_{\beta} $ assess misclassification probability, guiding selection of adversarial strategies to construct a minimax game and expand the detector's knowledge boundary.
Additionally, a regularization term ($-\lambda_4 \text{KL}(\Pi_t\|\Pi_{t-1})$) enforces policy consistency:
\begin{equation}
    \text{KL}(\Pi_t\|\Pi_{t-1}) = \mathbb{E}_{s\sim\mathcal{D}}[\Pi_t\log\frac{\Pi_t}{\Pi_{t-1}}]
\end{equation}
With $\lambda$ varying by training phase, overall reward function is:
\begin{equation}
\begin{aligned}
r_t &= \lambda_1 \mathbb{E}[1-\mathcal{C}_{tr}] + \lambda_2 \Delta \text{AUC}_{val} \\           & \quad +\lambda_3 \mathcal{C}_{adv} - \lambda_4 \text{KL}(\Pi_t|\Pi_{t-1})
\end{aligned}
\end{equation}
Given significant distributional differences among augmented forged samples, the environment is non-stationary. We use the PPO-Clip algorithm \cite{ppo} for policy optimization, which enforces hard clipping to limit policy updates (ensuring stability) and incorporates importance sampling with multiple updates to reduce sample demand and accelerate convergence.

\subsection{Multi-environment Causal Invariance Learning}
Guided by the causal theory, we create diverse environments based on the policy entropy and utilize Invariant Risk Minimization (IRM) to guide the detector in learning unbiased forgery features.

\subsubsection{Causal Analysis of Data Augmentation Bias.}
In deepfake detection, data augmentation essentially constitutes human intervention in the causal mechanisms of data generation. We model augmentation operators as intervention variables $A$, which are governed by the data construction strategy $M$(\(M \rightarrow A\)). From a causal perspective, features are categorized into causal features \(X_v\) and spurious features \(X_s\). The definitions of \(X_v\) and \(X_s\), along with the influence of $A$ on the feature space, are detailed in Appendix A.
\subsubsection{Policy Entropy-driven Environment Variation.}
Building on the above causal analysis, we propose a policy entropy-driven environment partitioning mechanism that explicitly distinguishes the decision confidence of different data augmentation strategies. This mechanism constructs training environments with significant conditional distribution discrepancies while preserving causally consistent cross-domain features. Confidence divergence in augmentation strategies across environments forces the model to decouple superficial correlations tied to specific tactics (e.g., blending-induced blur artifacts), enhancing its ability to capture environment-stable causal features. This environmental contrast fundamentally overcomes the feature disentanglement limitations of conventional augmentation. For a training batch with \( N \) samples, the environment construction process is as follows:
$\mathcal{H}_t = -\sum_{i=1}^N \sum_{j=1}^K \Pi_\phi(a_j|s_i)\log\Pi_\phi(a_j|s_i) $
where K denotes the number of augmentation strategies. The environment construction consists of two core parts:Dominant environment: Constructed by selecting samples with the absolute minimum entropy values across the entire batch to create a low-uncertainty environment:
$ \mathcal{E}_d^{(t)} = \{ x_{i^*} \mid i^* = \argmin_{1 \leq i \leq N} \mathcal{H}_t(i) \} $
Adversarial environment ensemble: Employing quantile partitioning to capture regions of high uncertainty.
\begin{equation}
\begin{aligned}
    \mathcal{E}_{\text{adv}}^1 &= \{ x_i \mid \mathcal{H}_t(i) \in [0.75,1.0] \cdot \mathcal{H}_t^{\max} \}, \\
    \mathcal{E}_{\text{adv}}^2 &= \{ x_i \mid \mathcal{H}_t(i) \in [0.50,0.75) \cdot \mathcal{H}_t^{\max} \}, \\
    \mathcal{E}_{\text{adv}}^3 &= \{ x_i \mid \mathcal{H}_t(i) \in [0.25,0.50) \cdot \mathcal{H}_t^{\max} \}.
\end{aligned}
\end{equation}



We also introduce a memory mechanism that maintains a first-in-first-out historical queue \(Q_m\)for each environment \( m \), with capacity equal to the batch size:
\begin{equation}
\begin{aligned}
    Q_m^{(t)} = \text{FIFO} & (Q_m^{(t-1)} \cup  \mathcal{E}_m^{(t)}C) \\
    \text{where} \quad \text{FIFO}(S, C) & \triangle  
\begin{cases}
     S, \text{if } |S| \leq C \\
     S_{|S|-C+1:|S|},  \text{if } |S| > C
\end{cases}
\end{aligned}
\end{equation}
Here, \( Q_m^{(t)} \) is the historical queue for the \( m \)-th environment at time step \( t \), \( \mathcal{E}_m^{(t)} \) is the set of samples selected at time step \( t \), and \( C \) is the capacity of the queue. The FIFO mechanism ensures that the queue's capacity does not exceed \( C \), and when the queue is full, the oldest sample is removed.

\subsubsection{Invariant Risk Minimization.}
\label{sec:IRM}
Building upon the aforementioned environment variation, we formulate an enhanced IRM objective that adapts to dynamically evolving training environments. The core innovation is the establishment of a bi-level optimization between environment-aware feature learning and augmentation policy adaptation:
\begin{equation}
\begin{aligned}
\min_\theta \sum_{m=1}^M &w_m \mathbb{E}{(x,y)\sim Q_m}[\mathcal{L}(f\theta(x),y)] \\
&+ \Omega |\nabla_{w|w=1}\mathcal{L}m(w \circ f\theta)|^2
\end{aligned}
\end{equation}
where \(\theta\) denotes the parameters of the shared backbone and domain bias classification heads. The backbone extracts features from input data, while the domain bias classification heads predict domain-specific biases. \(M\) represents the number of environments; \(Q_m\) stands for the historical queue of the \(m\)-th environment; \(\mathcal{L}\) is the loss function; and \(\Omega\) is the regularization weight that balances the main loss and gradient penalty. Additionally, \(w_m\) denotes the importance weight of environment \(m\), calculated as:
\begin{equation}
\begin{aligned}
    w_m &= \frac{\exp(-\bar{\mathcal{H}}_m)}{\sum_{k=1}^M \exp(-\bar{\mathcal{H}}_k)}, \\
    \text{where} \quad \bar{\mathcal{H}}_m &= \mathbb{E}_{x\sim Q_m}\left[\mathcal{H}(\Pi_\phi(a|x))\right].
\end{aligned}
\end{equation}
\subsection{Multi-dimensional Curriculum Scheduling}
\textbf{Data Course:} The proportion of augmented forged samples reflects the complexity of forgery features. In the early training stage, as the model has not yet mastered basic forgery identification, introducing a large number of advanced samples with complex features may slow down convergence due to the difficulty in processing such samples. Thus, we gradually incorporate augmented samples and design a sine-based proportion scheduling function.
\begin{equation}
q(t) = 0.5 + 0.5 \cdot \max(\min (\sin(\pi \cdot \frac{t - \tau/4}{\tau/2}), 1), -1)
\end{equation}
$\tau$ denotes total epochs, and t represents the current epoch.

\textbf{Exploration Course:} Entropy regularization incorporates an entropy regularization term into the reinforcement learning objective function to balance the exploration-exploitation trade-off in RL policies. Herein, a dynamic entropy regularization mechanism is proposed.
$\beta(t) = \beta_{\max} \cdot [\sigma(k(\frac{t}{\tau}-\mu)) - \sigma(-k\mu)] $
Here, \(\sigma\) is the sigmoid function, \(k=5\) controls the steepness of the transition, and \(\mu=0.3\) sets the peak phase of exploration intensity. The function maintains low exploration intensity in the early stage, linearly increases it in the middle stage to promote exploration of new strategies, and automatically anneals in later stage to stabilize the policy.

\textbf{Region Course:} We propose a facial curriculum template learning method based on dynamic region sampling. The method first constructs a region pool composed of 15 regions based on four basic facial organs (left eye, right eye, nose, mouth) and their combinations using facial landmark technology. It then dynamically controls the proportion of forgery region area using the exponential decay function:
$A(t) = A_{\text{full}} e^{-\lambda t} + A_{\text{min}} $
where \(A_{\text{full}} = 1.0\), \(A_{\text{min}} = 0.3\), and \(\lambda = 2/\tau\). During training, at each time step \(t\), candidate regions are selected from the region pool according to the current target area \(A(t)\), and then sampled randomly with Gaussian weights:
$p_i \propto \exp(-\frac{(A_i - A(t))^2}{2\sigma^2}) $
This allows the forgery region to gradually transition from multi-organ combinations in the early stages (\(t < 0.3\tau\), on average, 3.2 organs) to single-organ regions in the later stages (\(t \geq 0.7\tau\), accounting for 82\%).
\subsection{Loss Function}
The overall loss function comprises two key components, designed to optimize both classification performance and feature invariance:
$\mathcal{L}_{\text{total}} = \mathcal{L}_{\text{CE}} + \gamma \mathcal{L}_{\text{bias}} $.

\textbf{Cross-Entropy Loss ($\mathcal{L}_{\text{CE}}$):} 
As the primary classification loss, it is computed on augmented samples:
\begin{equation}
\begin{aligned}
\mathcal{L}_{\text{CE}} = -\frac{1}{N}\sum_{i=1}^N [y_i\log\sigma(f_\theta(x_i))   \\
  \quad+ (1-y_i)\log(1-\sigma(f_\theta(x_i)))]
\end{aligned}
\end{equation}
where $x_i$ denotes augmented samples, $y_i$ represents ground-truth labels, $f_\theta$ is the detector output and $\sigma$ stands for the sigmoid activation function.

\textbf{Bias Loss ($\mathcal{L}_{\text{bias}}$)}: This is the invariant learning regularization term derived from our enhanced IRM framework (see Section 'Invariant Risk Minimization' for derivation details). The hyperparameter \(\gamma\) balances the learning of discriminative features and the maintenance of feature invariance across different augmentation strategies.

\begin{table*}[h!]
\centering
\resizebox{0.96\linewidth}{!}{ 
\begin{tabular}
{l@{\hspace{0.24cm}}|@{\hspace{0.24cm}}c@{\hspace{0.24cm}}|@{\hspace{0.24cm}}c@{\hspace{0.24cm}}@{\hspace{0.24cm}}c@{\hspace{0.1cm}}c@{\hspace{0.1cm}}c@{\hspace{0.1cm}}c@{\hspace{0.1cm}}c@{\hspace{0.1cm}}c} \toprule[1.25pt]
\rule{0pt}{1.1em} 
Method    & Venues   &FF++   & CDFv1& CDFv2 & DFDC  & DFDCP & UADFV  & Avg. \\ \midrule[1.25pt]
Xception \cite{xception} &CVPR'17   &0.9637 &  0.7794     &    0.7365   &   0.7077  &   0.7374 & 0.9379   &  0.7798 \\
Meso4 \cite{mesonet}   & WIFS'18  & 0.6077  &   0.7358&       0.6091&       0.5560&     0.5994&   0.7150 & 0.6431\\
FWA \cite{FWA}   & CVPRW'18         & 0.8765& 0.7897 & 0.6680 & 0.6375 & 0.6132&   0.9049&         0.7433\\
EfficientB4 \cite{effnet} &ICML'19 & 0.9567 &    0.7909&       0.7487&       0.6955&     0.7283& 0.9472 &     0.7821\\
Capsule \cite{capsule}&ICASSP'19   & 0.8421  &   0.7909&       0.7472&       0.6465&     0.6568&  0.9078&0.7498\\
CNN-Aug \cite{cnn-Aug}   & CVPR'20& 0.8493 & 0.7420&       0.7027&      0.6361&     0.6170&   0.8739&  0.7143\\
X-ray \cite{xray}  & CVPR'20     & 0.9592 &   0.7093 & 0.6786 & 0.6326 & 0.6942&  0.8989&   0.7227\\
FFD \cite{ffd}   & CVPR'20      & 0.9624  &   0.7840 & 0.7435 & 0.7029 & 0.7426&  0.9450&         0.7836\\
F3Net \cite{F3Net} &  ECCV'20   & 0.9635  &   0.7769 & 0.7352 & 0.7021 & 0.7354&  0.9347    &   0.7769\\ 
SPSL \cite{spsl}   & CVPR'21    &0.9610  &  0.8150 & 0.7650 & 0.7040 & 0.7408 & 0.9424 &   0.7934\\ 
SRM \cite{srm}   &  CVPR'21     &  0.9576&  0.7926 & 0.7552 & 0.6995 & 0.7408 &  0.9427    &   0.7862\\ 
CORE \cite{core}   &  CVPRW'22   & 0.9638 &  0.7798 & 0.7428 & 0.7049 & 0.7341& 0.9412&        0.7726\\ 
Recce \cite{recce}  &  CVPR'22  & 0.9621 &  0.7677 & 0.7319 & 0.7133 & 0.7419&  0.9446 &       0.7794\\ 
SBI \cite{SBI}  &  CVPR'22       & 0.8176 & 0.8311 & 0.8015 & 0.7139 & 0.7794 & 0.9475& 0.8147  \\  
UCF \cite{ucf}     &  ICCV'23    & 0.9705&  0.7793 & 0.7527 &0.7191& 0.7594& 0.9528 &  0.7967\\ 
F-G \cite{f-g}     &  CVPR'24   & 0.9739 &  0.7330 & 0.7016 &0.6027& 0.6824& 0.8768 &  0.7193\\
LSDA \cite{lsda}     &  CVPR'24   & 0.9482 &  0.8467 & 0.8142 &0.7203& 0.7894& 0.9515 &  0.8244\\
Prodet \cite{Prodet}     &  NIPS'24   & 0.9591  &  0.8756 & 0.8420 &0.6973& 0.7744& 0.9539 &  0.8282\\
FreqBlender \cite{freqblender}     &  NIPS'24  & 0.8753  &  0.8928 & 0.8387 &0.7041& 0.7692& 0.9487 &  0.8307\\
\midrule[1.25pt]
\multirow{2}{*}{CRDA(ours)}  & \multirow{2}{*}{-}  &  \multirow{2}{*}{0.9374} &\textbf{0.9010} &\textbf{0.8536}&\textbf{0.7429}&\textbf{0.7973}&\textbf{0.9570}&\textbf{0.8504}\\ 
 &&&\small{({$\uparrow2.90\%$})}&\small{{($\uparrow1.38\%$)}}&\small{{($\uparrow3.14\%$)}}&\small{{($\uparrow1.00\%$)}}&\small{{($\uparrow0.03\%$)}}&\small{{($\uparrow2.68\%$)}}\\
\bottomrule[1.25pt]
\end{tabular}}
\caption{Cross-dataset evaluations (AUC) from FF++ \cite{FF++} (in-dataset) to CDFv1 \cite{Celeb-df}, CDFv2 \cite{Celeb-df}, DFDC \cite{DFDC-paper} , DFDCP \cite{DFDC-paper} and UADFV \cite{UADFV-paper} (cross-dataset). Avg. denotes the average value of cross-dataset results. The best results are highlighted in \textbf{bold}. Cross-dataset improvements compared with the previous best one are written in \small{small}.} \label{res-CRDA}
\end{table*}

\begin{table*}[h!]
\centering

\setlength{\tabcolsep}{0.6em} 
\begin{tabular}{l@{\hspace{0.7em}}c@{\hspace{0.7em}}c@{\hspace{0.7em}}c@{\hspace{0.7em}}c@{\hspace{0.7em}}c@{\hspace{0.7em}}c}
\toprule[1.25pt] 
\rule{0pt}{1.1em} 
\textbf{Backbone} & \textbf{CDFv1} & \textbf{CDFv2} & \textbf{DFDC} & \textbf{DFDCP} & \textbf{UADFV} & \textbf{Avg.} \\ 
\midrule[0.8pt] 

\rule{0pt}{1em} 
ResNet50 + SBI & 0.8438 & 0.8215 & 0.6862 & 0.7102 & 0.8958 & 0.7915 \\
ResNet50 + FreqBlender & 0.8527 & 0.8372 & 0.7016 & 0.7385 & 0.9124 & 0.8085 \\
ResNet50 + Ours & 0.8384 & 0.8301 & 0.6748 & 0.7159 & 0.8947 & 0.7908 \\[0.2em]
\midrule[0.5pt] 

Xception + SBI & 0.8427 & 0.8132 & 0.7069 & 0.7614 & 0.9318 & 0.8112 \\
Xception + FreqBlender & 0.8593 & 0.8261 & 0.7142 & 0.7716 & 0.9387 & 0.8220 \\ 
Xception + Ours & 0.8768 & 0.8425 & 0.7197 & 0.7859 & 0.9521 & 0.8354 \\[0.2em]
\midrule[0.5pt]

EfficientNet-B4 + SBI & 0.8311 & 0.8015 & 0.7139 & 0.7794 & 0.9475 & 0.8147 \\
EfficientNet-B4 + FreqBlender & 0.8928 & 0.8387 & 0.7041 & 0.7692 & 0.9487 & 0.8307 \\ 
EfficientNet-B4 + Ours & 0.9010 & 0.8536 & 0.7429 & 0.7973 & 0.9570 & 0.8504 \\[0.2em]
\midrule[0.5pt]

Swin-Transformer + SBI & 0.8793 & 0.8427 & 0.7196 & 0.7984 & 0.9347 & 0.8250 \\
Swin-Transformer + FreqBlender & 0.9012 & 0.8463 & 0.7248 & 0.8087 & 0.9395 & 0.8441 \\ 
Swin-Transformer + Ours & 0.9114 & 0.8488 & 0.7352 & 0.8264 & 0.9476 & 0.8539 \\
\bottomrule[1.25pt] 
\end{tabular}
\caption{Performance comparison across different backbone networks (AUC scores). The results show that our method can achieve performance improvements on various backbones except ResNet50 — highlights its adaptability.} 
\label{tab:backbone_comparison}
\end{table*} 

\section{Experiments}
\subsection{Experimental Setting.}
\textbf{Datasets.} Experiments use common deepfake datasets: FaceForensics++ (FF++)\cite{FF++}, Celeb-DF-v1/v2 (CDFv1/v2)\cite{Celeb-df}, DeepFake Detection Challenge (DFDC)/its preview (DFDCP)\cite{DFDC-paper}, and UADFV \cite{UADFV-paper}. FF++ has 1000 pristine and 4000 manipulated videos (4 techniques) with 3 compression levels; we use c23. CDF contains 590 pristine and 5639 fake videos. DFDC has 100,000 clips, DFDCP as its preview. UADFV includes 45 deepfake and 45 pristine videos. Original training/testing splits are adopted. \\
\textbf{Implementation Details.}
For preprocessing and training, we followed the official code and settings from \cite{deepfakebench} for a fair comparison. We used EfficientNetB4 \cite{effnet} as the detector's backbone. The PolicyNet consists of two convolutional layers and three MLPs with ReLU activation, using softmax to output a probability distribution over seven augmentation strategies. The domain head in the causal loss is identical to the detector's classifier. Reward coefficients ($\lambda$) were initialized to 0.6, 0.2, 0.1, and 0.1, and the causal regularization coefficient ($\gamma$) was 0.5. The detector was trained with Adam \cite{adam} (lr = $1 \times 10^{-4}$, weight decay = $5 \times 10^{-4}$, 30 epochs). The PPO \cite{ppo} algorithm used a learning rate of $3 \times 10^{-5}$, a GAE coefficient of 0.8, and a discount factor of 0.95. All experiments were conducted on eight NVIDIA Tesla V100 GPUs; see \textbf{Appendix C} for details.



\subsection{Overall Performance on Comprehensive Datasets}
To validate the effectiveness of our method, we trained it on the FF++ dataset and evaluated its performance on five other cross-domain datasets. In this study, we primarily report the area under the ROC curve (AUC) to facilitate comparisons with prior works. As shown in Table \ref{res-CRDA}, we conducted frame-level comparisons with 18 state-of-the-art methods (with detailed explanation provided in \textbf{Appendix B}), all evaluated under the same experimental settings as ours. EfficientB4 serves as the baseline using only deepfake data, while SBI \cite{SBI} and FreqBlender \cite{freqblender} are baselines using a single type of blendfake data. By progressively combining diverse forgery augmentation strategies, our method achieved the best performance across all datasets.

\begin{table*}[h!]
\centering
\resizebox{0.97\linewidth}{!}{ 
\begin{tabular}{llcccccc}
\toprule[1.25pt]
\rule{0pt}{1.0em} 
\textbf{Spatial-Domain} & \textbf{Freq-Domain} & \textbf{CDFv1} & \textbf{CDFv2} & \textbf{DFDC} & \textbf{DFDCP} & \textbf{UADFV} & \textbf{Avg.} \\
\midrule[0.8pt] 
 base\_spatial & FSBI & 0.8573 & 0.8451 & 0.7237 & 0.7428 & 0.9349 & 0.8209 \\
base\_spatial + SBI & FSBI & 0.8729 & 0.8493 & 0.7458 & 0.7846 & 0.9374 & 0.8380 \\
 base\_spatial + SBI & FreqBlender & 0.8842 & 0.8425 & 0.7393 & 0.8016 & 0.9443 & 0.8424 \\
 \rule{0pt}{1.0em} 
standard (both domains) & - & 0.9010 & 0.8536 & 0.7429 & 0.7973 & 0.9570 & 0.8504 \\
 base\_spatial+SBI+X-ray+PCL+I2G & FreqBlender+FSBI & 0.8937 & 0.8562 & 0.7258 & 0.8073 & 0.9615 & 0.8489 \\
\bottomrule[1.25pt]
\end{tabular}}
\caption{Ablation study on selection of augmentation strategy. More augmentations do not necessarily lead to better performance; future research should focus on improving the quality of augmentation schemes.}  
\label{tab:aug_ablation}
\end{table*}

\begin{table}[h!]
\centering
\setlength{\tabcolsep}{0.5em} 
\begin{tabular}{l@{\hspace{0.5em}}c@{\hspace{0.5em}}c@{\hspace{0.5em}}c}
\toprule[1.25pt] 
\rule{0pt}{1.0em} 
\textbf{Method} & \textbf{CDFv2} & \textbf{DFDC} \\ 
\midrule[0.8pt] 

\rule{0pt}{0.1em} 
Two-branch (ECCV'2020) & 76.7 & -- \\
Face X-ray (CVPR'2020) & 79.5 & 65.5 \\
LipForensics (CVPR'2021) & 82.4 & 73.5 \\
PCL+I2G (ICCV'2021) & 90.0 & 67.5 \\
FTCN (ICCV'2021) & 86.9 & 74.0 \\
HCIL (ECCV'2022) & 79.0 & 69.2 \\
AUNet (CVPR'2023) & 92.8 & 73.8 \\
AltFreezing (CVPR'2023) & 89.5 & -- \\[0.2em]
\midrule[0.5pt] 

\textbf{EffcientNetB4 + Ours} & \textbf{95.22} & \textbf{82.64} \\
\bottomrule[1.25pt] 
\end{tabular}
\caption{Performance comparison on video-level. CRDA surpasses specialized approaches by a margin of over 5\%.}
\label{tab:video_performance_comparison}
\end{table}

\begin{table}[h]
    \centering
    \renewcommand{\arraystretch}{1.2} 
    \begin{tabular}{c ccc cc}
    \toprule[1.25pt] 
    ID & \multicolumn{3}{c}{Components} & \multicolumn{2}{c}{AUC(\%)} \\ 
    \cline{2-4} \cline{5-6}
    & RL-DA & CIL & CS & CelebDF-v2 & DFDC \\ 
    \midrule[0.8pt] 
    \rule{0pt}{0em}
    1 & × & × & × & 74.87  & 69.55  \\
    2 & × & × & \checkmark & 76.34  & 70.22 \\
    3 & \checkmark & × & \checkmark & 81.37  & 72.81 \\
    4 & × & \checkmark & \checkmark  & 80.95  & 73.16\\
    5 & \checkmark & \checkmark & \checkmark & 85.36 & 74.29 \\
    \bottomrule[1.25pt] 
    \end{tabular}
    \caption{Ablation study. 'RL-DA' refers to our RL-based data augmentations. 'CIL' refers to Multi-environment Causal Invariance Learning. 'CS' refers to Curriculum Scheduling.}
    \label{tab:ablation}
\end{table}

\begin{figure}
    \centering
    \includegraphics[width=0.88\linewidth]{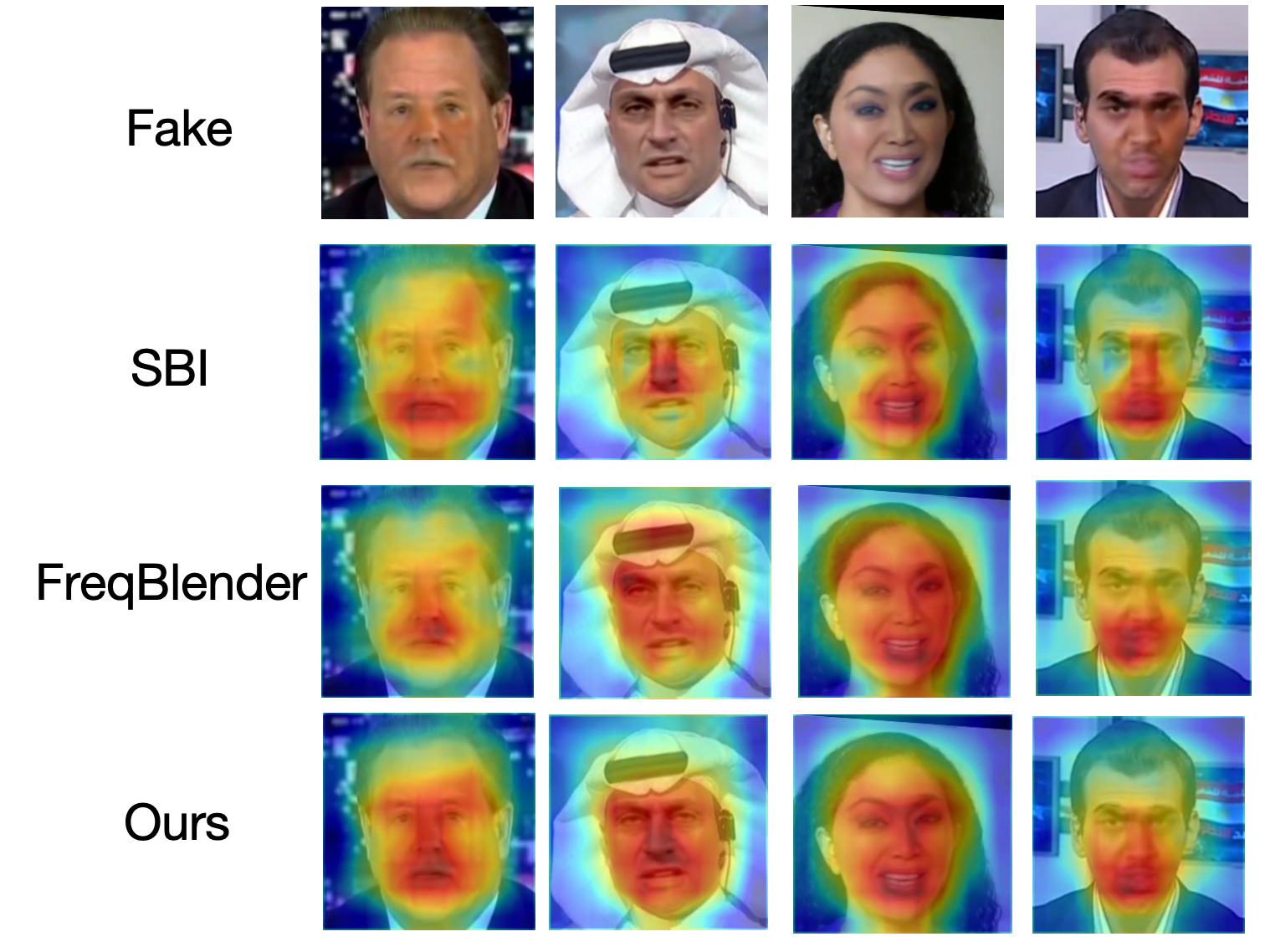}
    \caption{Grad-CAM visualization of SBI, FreqBlender and CRDA. Compared with baselines, CRDA focuses more on the manipulated structural boundaries.}
    \label{fig:gradcam}
\end{figure}

\subsection{Ablation Study On CRDA's Generalizability}
\textbf{Results in video-level:}
Table ~\ref{tab:video_performance_comparison} shows the video-level results obtained by averaging the prediction results of each frame in the evaluation video, and evaluates them using the AUC metric. Compared with SOTA video-level methods, our method achieves a relative improvement of above 5\% in cross-dataset generalization, demonstrating its superior capacity in temporal generalization while maintaining frame-level performance. \\
\textbf{Effect on different backbones:} We evaluate CRDA on various backbones (ResNet-50, Xception, EfficientNet-B4, Swin-Transformer) against SBI and FreqBlender. The results in Table~\ref{tab:backbone_comparison} confirm its model-agnostic nature, with consistent gains on Xception (+1.34\%), EfficientNet-B4 (+1.97\%), and Swin-Transformer (+1.22\%). The minor performance loss on ResNet-50 is likely because its weaker feature extraction limits the effectiveness of our method’s operations.\\
\textbf{Selection of augmentation strategy:} We analyze various spatial and frequency-domain augmentations as RL actions (Table~\ref{tab:aug_ablation}). Results show that increasing the quantity of augmentations beyond a standard set yields negligible performance gains. This highlights that our method’s performance is tied to the quality of the augmentation scheme, not its quantity. Consequently, our method is scalable and will benefit from future advancements in higher-quality augmentations. \\
\textbf{Saliency Visualization:} We employ Grad-CAM \cite{grad-cam++} to visualize the attention of our method compared to SBI and FreqBlender on four manipulations in the FF++ dataset. CRDA pays less attention to irrelevant regions, with enhanced focus on facial regions.

\subsection{Ablation Study On Key Componments}
To better understand the contribution of each module, we conduct an ablation study on CRDA by decomposing it into three components: RL-DA (Reinforcement Learning-based Data Augmentation), CIL (Multi-environment Causal Invariance Learning), and CS (Curriculum Scheduling). The experimental setup is detailed in Table \ref{tab:ablation}. While the CS module contributes the least individually, its combination with other modules significantly boosts performance. Ultimately, the synergistic effect of all three components yields the best results. Additionally, we conducted ablation studies on the choice of RL algorithm and its specific parameter settings, which are detailed in the \textbf{Appendix D}.

\section{Conclusion}
In this work, we introduce CRDA: an RL-driven data augmentation strategy that significantly improves detector performance by generating a diversified training environment. A key trade-off of CRDA is the increased computational cost, and a limitation is the lack of a quantitative method to guide the selection of base augmentation strategies. Our future work will address these issues by optimizing efficiency and establishing a robust evaluation framework.

\onecolumn  
\newpage    
\twocolumn  
\bibliography{aaai2026}

\onecolumn  
\newpage    
\twocolumn  

\end{document}